\pdfoutput=1
\documentclass[a4paper,11pt]{article}
\usepackage{eacl2012}
\makeatletter
\newcommand{\@BIBLABEL}{\@emptybiblabel}
\newcommand{\@emptybiblabel}[1]{}
\makeatother
\usepackage{times}
\usepackage{latexsym}
\usepackage{amsmath}
\usepackage{multirow}
\usepackage{url}
\usepackage[pdftex]{graphicx}
\usepackage{subfig}
\usepackage{sidecap}
\usepackage{setspace}
\usepackage{morefloats}
\usepackage{wrapfig}
\usepackage{fancyhdr} 
\usepackage{hyperref}

\title{A random forest system combination approach for error detection in digital dictionaries}

\author{Michael Bloodgood \and Peng Ye \and Paul Rodrigues \\
and {\bf David Zajic} and {\bf David Doermann} \\
University of Maryland \\
College Park, MD \\
{\tt meb@umd.edu, pengye@umiacs.umd.edu, prr@umd.edu,}\\ {\tt dzajic@casl.umd.edu, doermann@umiacs.umd.edu} \\
}

\date{}

\begin{document}

\thispagestyle{fancy}

\maketitle
\begin{abstract}

When digitizing a print bilingual dictionary, whether via optical character recognition or manual entry, 
it is inevitable that errors are introduced into the electronic version that is created. 
We investigate automating the process of detecting errors in an XML representation of a digitized print dictionary using a 
hybrid approach that combines rule-based, feature-based, and 
language model-based methods. 
We investigate combining methods and show that using random forests is a promising approach.
We find that in isolation, unsupervised methods rival the performance of supervised methods. 
Random forests typically require training 
data so we investigate how we can apply random forests to combine individual base methods that are themselves unsupervised 
without requiring large amounts of 
training data. Experiments reveal empirically that a relatively small amount of data is sufficient
and can potentially be further reduced through specific selection criteria.

\end{abstract}

\section{Introduction} \label{introduction}

\begin{figure}
  \centering
  \includegraphics[scale=0.85]{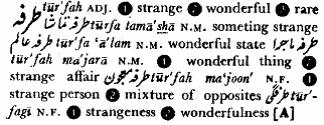}
  \caption{Example dictionary entry} 
  \label{fig:exampleErrorPrintScan}
\end{figure}

\begin{figure}
  \centering
  \includegraphics[scale=0.50]{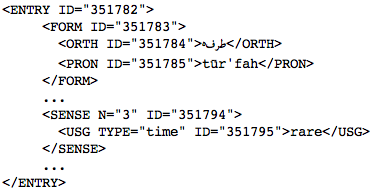}
  \caption{Example of error in XML}  
  \label{fig:exampleErrorXMLBefore}
\end{figure}

\begin{figure}
  \centering
  \includegraphics[scale=1.5]{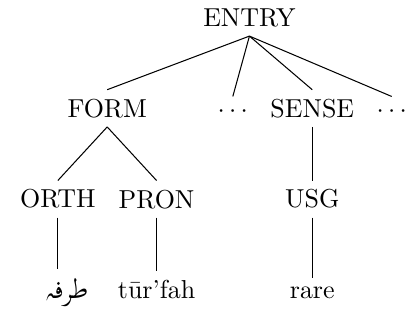}
  \caption{Tree structure of error}
  \label{fig:exampleErrorTreeBefore} 
\end{figure}

\begin{figure}
  \centering
  \includegraphics[scale=0.6]{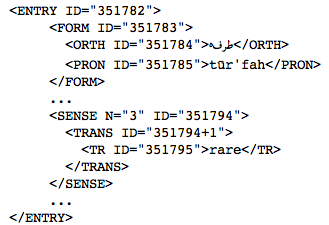}
  \caption{Example of error in XML, fixed}
  \label{fig:exampleErrorXMLAfter} 
\end{figure}

\begin{figure}
  \centering
  \includegraphics[scale=1.5]{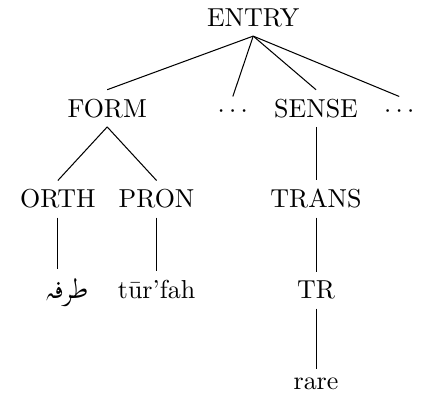}
  \caption{Tree structure of error, fixed}
  \label{fig:exampleErrorTreeAfter} 
\end{figure}

Digital versions of bilingual dictionaries often have errors that need to be fixed. For example, 
Figures \ref{fig:exampleErrorPrintScan} through \ref{fig:exampleErrorTreeAfter} show an example 
of an error that occurred in one of our development dictionaries and how the error should be corrected. 
Figure~\ref{fig:exampleErrorPrintScan} shows 
the entry for the word 
``turfah'' as it appeared in the original print copy of \cite{qureshi1991}. 
We see this word has three senses with slightly different meanings. The third sense is ``rare''. 
In the original digitized XML version of \cite{qureshi1991} depicted in Figure~\ref{fig:exampleErrorXMLBefore}, this was 
misrepresented as not being the meaning of ``turfah'' but instead being a usage note that frequency of use 
of the third sense was rare. Figure~\ref{fig:exampleErrorTreeBefore} shows the tree corresponding to this XML representation. 
The corrected digital XML representation is depicted in Figure~\ref{fig:exampleErrorXMLAfter} and the corresponding 
corrected tree is shown in Figure~\ref{fig:exampleErrorTreeAfter}.

\newcite{zajic2011} presented a method for repairing a digital dictionary in an XML format using
a dictionary markup language called DML. It remains time-consuming and error-prone however to have a human read through and manually correct a digital 
version of a dictionary, even with languages such as DML available. We therefore investigate automating the detection of errors. 

We investigate the use of three individual methods. 
The first is a supervised feature-based method trained using 
SVMs (Support Vector Machines). The second is a language-modeling method that replicates the method presented in \cite{rodrigues2011}. 
The third is a simple rule
inference method. The three individual methods have different performances. So we investigate how we can combine 
the methods most effectively. We experiment with majority vote, score combination, and random forest methods and find that random forest combinations work the best. 

For many dictionaries, training data will not be available in large quantities a priori and therefore methods that require only small 
amounts of training data are desirable. 
Interestingly, for automatically detecting errors in dictionaries, we find that the unsupervised methods have performance that rivals that of the 
supervised feature-based method trained using SVMs. 
Moreover, when we combine methods using the random forest method, the combination of unsupervised methods works better
than the supervised method in isolation and almost as well as the combination of all available methods. 
A potential drawback of using the random forest combination 
method however is that it requires training data. 
We investigated how much training data is needed and find that the amount of training data required is modest.
Furthermore, by selecting the training data to be labeled with the use of specific selection methods reminiscent of active learning, 
it may be possible to train the random forest system combination method with even less data without sacrificing performance. 

In section~\ref{relatedWork} we discuss previous related work and in section~\ref{singleMethods} we explain the 
three individual methods we use for our application.
In section~\ref{combinedMethods} we explain the three methods we explored for combining 
methods; in section~\ref{experiments} we present and discuss experimental results
and in section~\ref{conclusions} we conclude and discuss future work.

\section{Related Work} \label{relatedWork}

Classifier combination techniques can be broadly classified into two categories: mathematical and behavioral \cite{tulyakov08}. 
In the first category, functions or rules combine normalized classifier scores from individual classifiers. 
Examples of techniques in this category include Majority Voting \cite{Lam97}, as well as simple score combination rules such as: sum rule, min rule, max rule and 
product rule \cite{Kittler98,Ross03,jain05}. In the second category, the output of 
individual classifiers are combined to form a feature vector as the input to a generic classifier such 
as classification trees \cite{Verlinde99,Ross03} or the k-nearest neighbors classifier \cite{Verlinde99}. 
Our method falls into the second category, where we use a random forest for system combination. 

The random forest method is described in \cite{breiman01}. 
It is an ensemble classifier consisting of a collection of decision trees (called a random forest) and the output of the random forest is the 
mode of the classes output by the individual trees. 
Each single tree is trained as follows: 1) a random set of samples from the initial training set is selected as a training set 
and 2) at each node of the tree, a random subset of the features is selected, and the locally optimal 
split is based on only this feature subset. The tree is fully grown without pruning. 
\newcite{ma05} used random forests for combining scores of 
several biometric devices for identity verification and have shown encouraging results. They use all fully supervised methods.
In contrast, we explore minimizing the amount of training data needed to train a random forest of unsupervised methods.

The use of active learning in order to reduce training data requirements without sacrificing model performance 
has been reported on extensively in the literature (e.g., \cite{seung1992,cohn1994,lewis1994,cohn1996,freund1997}).
When training our random forest combination of individual methods that are themselves unsupervised, we explore 
how to select the data so that only small amounts of training data are needed because for 
many dictionaries, gathering training data may be expensive and labor-intensive.

\section{Three Single Method Approaches for Error Detection} \label{singleMethods}

Before we discuss our approaches for combining systems, we briefly explain the three individual 
systems that form the foundation of our combined system.

First, we use a supervised approach where we train a model using SVM$^{light}$ \cite{joachims1999} with a linear kernel and default regularization parameters.
We use a depth first traversal of the XML tree and use unigrams and bigrams of the tags that occur as features for each subtree to make a classification decision.

We also explore two unsupervised approaches. 
The first unsupervised approach learns rules for when to classify nodes as errors or not.
The rule-based method computes an anomaly score based on the probability of subtree structures. 
Given a structure A and its probability P(A), the event that A occurs has anomaly score 1-P(A) and the event that A does not occur has anomaly score P(A). 
The basic idea is if a certain structure happens rarely, i.e. P(A) is very small, then the occurrence of A should have a high anomaly score. 
On the other hand, if A occurs frequently, then the absence of A indicates anomaly. 
To obtain the anomaly score of a tree, we simply take the maximal scores of all events induced by subtrees within this tree. 

The second unsupervised approach uses a reimplementation of the language modeling method described in \cite{rodrigues2011}. 
Briefly, this methods works by calculating the probability a flattened XML branch can occur, given a 
probability model trained on the XML branches from the original dictionary.  We used \cite{stolcke:2002} to 
generate bigram models using Good Turing smoothing and Katz back off, and evaluated the log probability of the 
XML branches, ranking the likelihood.  The first 1000 branches were submitted to the hybrid system marked as an error, 
and the remaining were submitted as a non-error. Results for the individual classifiers are presented in 
section~\ref{experiments}.

\section{Three Methods for Combining Systems} \label{combinedMethods}

We investigate three methods for combining the three individual methods. 
As a baseline, we investigate simple majority vote.
This method takes the classification decisions of the three methods and assigns the final classification as the classification that 
the majority of the methods predicted.

A drawback of majority vote is that it does not weight the votes at all. However, it might make sense to weight the votes according to factors such 
as the strength of the classification score. For example, all of our classifiers make binary decisions but output scores that are indicative of the  confidence of their classifications. Therefore we also explore a score combination method that considers these scores. 
Since measures from the different systems are in different ranges, we normalize these measurements before 
combining them \cite{jain05}. We use z-score which computes the arithmetic mean and standard deviation of the given 
data for score normalization. We then take the summation of normalized measures as the final measure. 
Classification is performed by thresholding this final measure.\footnote{In our experiments we used 0 as the threshold.}

Another approach would be to weight them by the performance level of the various constituent classifiers in the ensemble. 
Weighting based on performance level of the individual classifiers is difficult because it would require extra labeled data to estimate the various performance levels. It is not clear how to translate the different performance estimates into weights, or how to have those weights interact with weights based 
on strengths of classification. Therefore, we did not weigh based on performance level explicitly.

We believe that our third combination method, the use of random forests, implicitly captures weighting based on performance level and strengths of
classifications. Our random forest approach uses three features, one for each of the individual systems we use. 
With random forests, strengths of classification are taken into account because they form the values of the three features we use. 
In addition, the performance level is taken into account because the training data used to train the decision trees 
that form the forest help to guide binning of the
feature values into appropriate ranges where classification decisions are made correctly. 
This will be discussed further in section~\ref{experiments}. 

\section{Experiments} \label{experiments}

This section explains the details of the experiments we conducted testing the performance of the various individual and combined systems.
Subsection~\ref{setup} explains the details of the data we experiment on; subsection~\ref{results} provides a summary of 
the main results of our experiments;
and subsection~\ref{discussion} discusses the results.

\subsection{Experimental Setup} \label{setup}

We obtained the data for our experiments using a digitized version of \cite{qureshi1991}, the same Urdu-English dictionary that \newcite{zajic2011} had used. 
\newcite{zajic2011} presented DML, a programming language used to fix errors in XML documents that contain lexicographic data.
A team of language experts used DML to correct errors in a digital, XML representation of the Kitabistan Urdu dictionary.
The current research compared the source XML document and the DML commands to identify the elements that the language experts decided to modify. 
We consider those elements to be errors. This is the ground truth used for training and evaluation.
We evaluate at two tiers, corresponding to two node types in the XML representation of the dictionary: ENTRY and SENSE.
The example depicted in Figures \ref{fig:exampleErrorPrintScan} through \ref{fig:exampleErrorTreeAfter} 
shows an example of SENSE. 
The intuition of the tier is that errors are detectable (or learnable) from observing the elements within a tier, and do not cross tier boundaries. 
These tiers are specific to the Kitabistan Urdu dictionary, and we selected them by observing the data. 
A limitation of our work is that we do not know at this time whether they are generally useful across dictionaries. 
Future work will be to automatically discover the meaningful evaluation tiers for a new dictionary.
  After this process, we have a dataset with 15,808 Entries, of which 47.53\% are marked as errors and 78,919 Senses, of which 10.79\% are marked as errors. 
We perform tenfold cross-validation in all experiments. In our random forest experiments, we use 12 decision trees, each with only 1
feature.

\subsection{Results} \label{results}

This section presents experimental results, first for individual systems and then for combined systems. 

\subsubsection{Performance of individual systems}
Tables \ref{tab:indi_entry} and \ref{tab:indi_sense} show the performance of language modeling-based method (LM), rule-based 
method (RULE) and the supervised feature-based method (FV) at different tiers. 
As can be seen, at the ENTRY tier, RULE obtains the highest F1-Measure and accuracy, while at the SENSE tier, FV performs 
the best. 

\begin{table}\footnotesize
\begin{center}
\begin{tabular}{|c|c|c|c|c|}
\hline
    &  Recall & Precision & F1-Measure & Accuracy \\
\hline
LM   & 11.97 & 89.90 & 21.13 & 57.53  \\
RULE & 99.79 & 70.83  & 82.85 & 80.37  \\
FV   & 35.34 & 93.68  & 51.32 & 68.14  \\
\hline
\end{tabular}
\end{center}
\caption{Performance of individual systems at ENTRY tier.}
\label{tab:indi_entry}
\end{table}

\begin{table}\footnotesize
\begin{center}
\begin{tabular}{|c|c|c|c|c|}
\hline
   &  Recall & Precision & F1-Measure & Accuracy \\
\hline
LM	& 9.85 & 94.00 & 17.83 & 90.20	\\
RULE & 84.59 & 58.86 & 69.42 & 91.96\\
FV & 72.44 & 98.66 & 83.54 & 96.92\\
\hline
\end{tabular}
\end{center}
\caption{Performance of individual systems at SENSE tier.}
\label{tab:indi_sense}
\end{table}

\subsubsection{Improving individual systems using random forests}
In this section, we show that by applying random forests on top of the output of individual systems, we can have 
gains (absolute gains, not relative) in accuracy of 4.34\% to 6.39\% and gains (again absolute, not relative) in F1-measure of 
3.64\% to 11.39\%. 
Tables \ref{tab:indi_rulerf_entry} and \ref{tab:indi_rulerf_sense} show our experimental results at ENTRY and SENSE tiers when applying random forests with 
the rule-based method.\footnote{We also applied random forests to our language modeling and feature-based methods, and saw similar gains in performance.}
These results are all obtained from 100 iterations of the experiments with different partitions of the training data chosen at each iteration.
Mean values of different evaluation measures and their standard deviations are shown in these tables. 
We change the percentage of training data and repeat the experiments to see how the amount of training data affects performance. 

\begin{figure}
  \centering
  \includegraphics[width=0.48\textwidth]{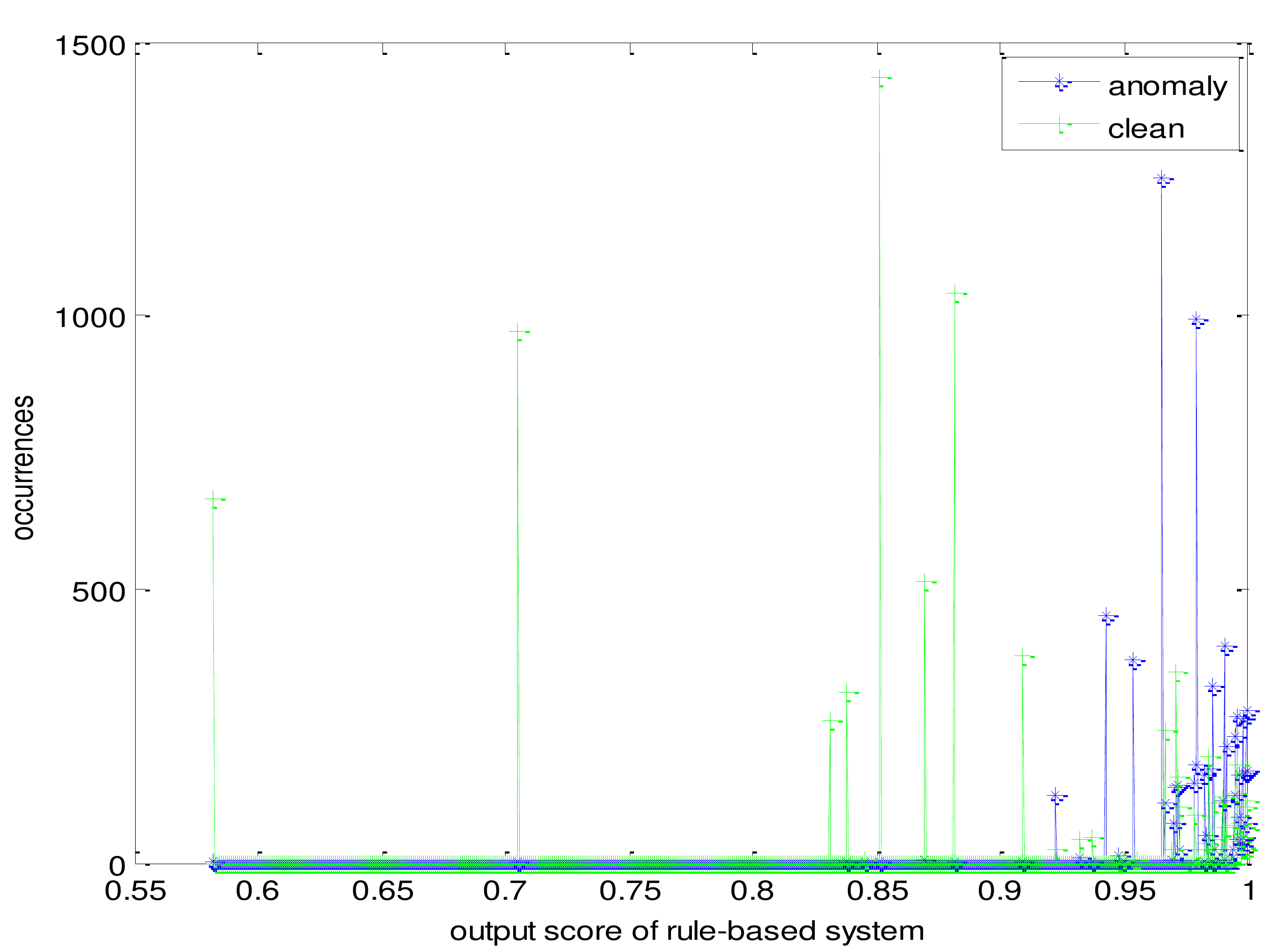}
  \caption{Output anomalies score from RULE (ENTRY tier).}
  \label{fig:rule_output}
\end{figure}

It might be surprising to see the gains in performance that can be achieved by using a random forest of decision trees created using only the rule-based scores
as features. 
To shed light on why this is so, we show the distribution of RULE-based output scores for anomaly nodes and clean nodes in Figure \ref{fig:rule_output}. 
They are well separated and this explains why RULE alone can have good performance. 
Recall RULE classifies nodes with anomaly scores larger than 0.9 as errors.
However, in Figure~\ref{fig:rule_output}, we can see that there are many clean
nodes with anomaly scores larger than 0.9. Thus, the simple 
thresholding strategy will bring in errors.
Applying random forest will help us identify these errorful regions to improve the performance. 
Another method for helping to identify these errorful regions and classify them correctly is to apply random forest of RULE combined with the other methods,
which we will see will even further boost the performance.

\begin{table*}\footnotesize
\begin{center}
\begin{tabular}{|c|c|c|c|c|}
\hline
Training \%  &  Recall & Precision & F1-Measure & Accuracy \\
\hline
0.1 &  78.17( 14.83) & 75.87( 3.96) & 76.18( 7.99) & 77.68( 5.11) \\				
1 &  82.46( 4.81) & 81.34( 2.14) & 81.79( 2.20) & 82.61( 1.69) \\				
10 &  87.30( 1.96) & 84.11( 1.29) & 85.64( 0.46) & 86.10( 0.35) \\				
50 &  89.19( 1.75) & 83.99( 1.20) & 86.49( 0.34) & 86.76( 0.28) \\			
\hline
\end{tabular}
\end{center}
\caption{Mean and std of evaluation measures from 100 iterations of experiments using RULE+RF. (ENTRY tier)}
\label{tab:indi_rulerf_entry}
\end{table*}
	
\begin{table*}\footnotesize
\begin{center}
\begin{tabular}{|c|c|c|c|c|}
\hline
Training \%  &  Recall & Precision & F1-Measure & Accuracy \\
\hline
0.1 &  60.22( 12.95) & 69.66( 9.54) & 63.29( 7.92) & 92.61( 1.57) \\				
1 &  70.28( 3.48) & 86.26( 3.69) & 77.31( 1.39) & 95.55( 0.25) \\				
10 &  71.52( 1.23) & 91.26( 1.39) & 80.18( 0.41) & 96.18( 0.07) \\				
50 &  72.11( 0.75) & 91.90( 0.64) & 80.81( 0.39) & 96.30( 0.06) \\				
\hline
\end{tabular}
\end{center}
\caption{Mean and std of evaluation measures from 100 iterations of experiments using RULE+RF. (SENSE tier)}
\label{tab:indi_rulerf_sense}
\end{table*}

\subsubsection{System combination}

In this section, we explore different methods for combining measures from the three systems. 
Table \ref{tab:comb1} shows the results of majority voting and score combination at the ENTRY tier. 
As can be seen, majority voting performs poorly. This may be due to the fact that the performances of the three systems are very different. 
RULE significantly outperforms the other two systems, and as discussed in Section~\ref{combinedMethods} neither majority voting 
nor score combination weights this higher performance appropriately.

\begin{table*}\footnotesize
\begin{center}
\begin{tabular}{|c|c|c|c|c|}
\hline
Method &  Recall & Precision & F1-Measure & Accuracy \\
\hline
Majority voting & 36.71 & 90.90 & 52.30 & 68.18 \\
Score combination & 76.48 &  75.82 & 76.15 & 77.23	\\
\hline
\end{tabular}
\end{center}
\caption{LM+RULE+FV (ENTRY tier)}
\label{tab:comb1}
\end{table*}

Tables \ref{tab:comb2_entry} and \ref{tab:comb2_sense} show the results of combining RULE and LM. 
This is of particular interest since these two systems are unsupervised. 
Combining these two unsupervised systems works better than the individual methods, including supervised methods. 
Tables \ref{tab:comb3_entry} and \ref{tab:comb3_sense} show the results for combinations of all available systems. 
This yields the highest performance, but only slightly higher than the combination of only unsupervised base methods.  

The random forest combination technique does require labeled data even if the underlying base methods are unsupervised. 
Based on the observation in Figure~\ref{fig:rule_output}, we further study whether choosing more 
training data from the most errorful regions will help to improve the performance. 
Experimental results in Table \ref{tab:exp2} show how the choice of training data affects performance. 
It appears that there may be a weak trend toward higher performance 
when we force the selection of the majority of the training data to be from 
ENTRY nodes whose RULE anomaly scores are larger than 0.9. However, the magnitudes of the observed differences in performance 
are within a single standard deviation so it remains for future work to determine if there are ways to select the training data
for our random forest combination in ways that substantially improve upon random selection.

\begin{table*}\footnotesize
\begin{center}
\begin{tabular}{|c|c|c|c|c|}
\hline
Training \% &  Recall & Precision & F1-Measure & Accuracy \\
\hline
0.1 &  77.43( 15.14) & 72.77( 6.03) & 74.26( 8.68) & 75.32( 6.71) \\				
1 &  86.50( 3.59) & 80.41( 1.95) & 83.27( 1.33) & 83.51( 1.11) \\				
10 &  88.12( 1.12) & 84.65( 0.57) & 86.34( 0.46) & 86.76( 0.39) \\				
50 &  89.12( 0.62) & 87.39( 0.56) & 88.25( 0.30) & 88.72( 0.29) \\					
\hline
\end{tabular}
\end{center}
\caption{System combination based on random forest (LM+RULE). (ENTRY tier, mean (std))}
\label{tab:comb2_entry}
\end{table*}

\begin{table*}\footnotesize
\begin{center}
\begin{tabular}{|c|c|c|c|c|}
\hline
Training \% &  Recall & Precision & F1-Measure & Accuracy \\
\hline
0.1 &  65.85( 12.70) & 71.96( 7.63) & 67.68( 7.06) & 93.38( 1.03) \\				
1 &  80.29( 3.58) & 84.97( 3.13) & 82.45( 1.36) & 96.31( 0.28) \\				
10 &  82.68( 2.49) & 90.91( 2.37) & 86.53( 0.41) & 97.22( 0.07) \\				
50 &  83.22( 2.43) & 92.21( 2.29) & 87.42( 0.35) & 97.42( 0.04) \\					
\hline
\end{tabular}
\end{center}
\caption{System combination based on random forest (LM+RULE). (SENSE tier, mean (std))}
\label{tab:comb2_sense}
\end{table*}

\begin{table*}\footnotesize
\begin{center}
\begin{tabular}{|c|c|c|c|c|}
\hline
Training \% &  Recall & Precision & F1-Measure & Accuracy \\
\hline
20 &  91.57( 0.55) & 87.77( 0.43) & 89.63( 0.23) & 89.93( 0.22) \\				
50 &  92.04( 0.54) & 88.85( 0.48) & 90.41( 0.29) & 90.72( 0.28) \\				
\hline
\end{tabular}
\end{center}
\caption{System combination based on random forest (LM+RULE+FV). (ENTRY tier, mean (std))}
\label{tab:comb3_entry}
\end{table*}

\begin{table*}\footnotesize
\begin{center}
\begin{tabular}{|c|c|c|c|c|}
\hline
Training \% &  Recall & Precision & F1-Measure & Accuracy \\
\hline
20 &  86.47( 1.01) & 90.67( 1.02) & 88.51( 0.26) & 97.58( 0.06) \\				
50 &  86.50( 0.81) & 92.04( 0.85) & 89.18( 0.30) & 97.73( 0.06) \\				
\hline
\end{tabular}
\end{center}
\caption{System combination based on random forest (LM+RULE+FV). (SENSE tier, mean (std))}
\label{tab:comb3_sense}
\end{table*}

\begin{table*}\footnotesize
\begin{center}
\begin{tabular}{|c|c|c|c|c|}
\hline
     &  Recall & Precision & F1-Measure & Accuracy \\
\hline
50\% &  85.40( 4.65) & 80.71( 3.49) & 82.82( 1.57) & 82.63( 1.54) \\
70\% &  86.13( 3.94) & 80.97( 2.64) & 83.36( 1.33) & 83.30( 1.21) \\
90\% &  85.77( 3.61) & 81.82( 2.72) & 83.65( 1.45) & 83.69( 1.35) \\
95\% &  85.93( 3.46) & 82.14( 2.98) & 83.89( 1.32) & 83.94( 1.18) \\
random &  86.50( 3.59) & 80.41( 1.95) & 83.27( 1.33) & 83.51( 1.11) \\	
\hline
\end{tabular}
\end{center}
\caption{Effect of choice of training data based on rule based method (Mean evaluation measures from 100 iterations of experiments using RULE+LM at ENTRY tier). 
We choose 1\% of the data for training and the first column in the table specifies the percentage of training data chosen 
from Entries with anomalous score larger than 0.9.}
\label{tab:exp2}
\end{table*}

\subsection{Discussion} \label{discussion}
Majority voting (at the entry level) performs poorly, since the performance of the three individual systems are 
very different and majority voting does not weight votes at all.
Score combination is a type of weighted voting. It takes into account the confidence level of output 
from different systems, which enables it to perform better than majority voting. 
However, score combination does not take into account the performance levels of the different 
systems, and we believe this limits its performance compared with random forest combinations. 

Random forest combinations perform the best, but the cost is that it is a supervised combination method. 
We investigated how the amount of training data affects the performance, and found that a small amount of labeled data 
is all that the random forest needs in order to be successful. Moreover, although this requires further exploration, 
there is weak evidence that the size of the labeled 
data can potentially be reduced by choosing it carefully from the region that is expected to be most errorful. 
For our application with a rule-based system, this is the
high-anomaly scoring region because although it is true that anomalies are often errors, it is 
also the case that some structures occur rarely but are not errorful. 

RULE+LM with random forest is a little better than RULE with random forest, with gain of about 0.7\% on F1-measure 
when evaluated at the ENTRY level using 10\% data for training.

\begin{figure}
  \centering
  \includegraphics[scale=0.6]{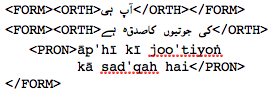}
  \caption{Example of error in XML}  
  \label{fig:zajic2}
\end{figure}

An examination of examples that are marked as being errors in our ground truth but that were not detected to be errors
by any of our systems suggests that some examples are decided on the basis of features
not yet considered by any system. For example, in Figure~\ref{fig:zajic2} the second FORM is well-formed structurally, but 
the Urdu text in the first FORM is the beginning of the phrase transliterated in the second FORM. Automatic systems
detected that the first FORM was an error, however did not mark the second FORM as an error whereas our ground truth marked
both as errors. 

Examination of false negatives also revealed cases where the systems were correct that there was no error
but our ground truth wrongly indicated that there was an error. These were due 
to our semi-automated method for producing
ground truth that considers elements mentioned in DML commands to be errors. We discovered instances in which merely 
mentioning an
element in a DML command does not imply that the element is an error. These cases are useful for 
making refinements to how ground truth is generated from DML commands. 

Examination of false positives revealed two categories. One was where the element is indeed an error but was not marked as
an element in our ground truth because it was part of a larger error that got deleted and therefore no DML command ever 
mentioned the smaller element but lexicographers upon inspection agree that the smaller element is indeed errorful. The
other category was where there were actual errors that the dictionary editors didn't repair with DML but that should have 
been repaired. 

A major limitation of our work is testing how well it generalizes to detecting errors in other dictionaries besides the Urdu-English one \cite{qureshi1991} that we conducted
our experiments on. 

\section{Conclusions} \label{conclusions}
We explored hybrid approaches for the application of automatically detecting 
errors in digitized copies of dictionaries. 
The base methods we explored consisted of a variety of unsupervised and
supervised methods. 
The combination methods we explored also consisted of some methods which
required labeled data and some which did not. 

We found that our base methods had different levels of performance and with this
scenario majority voting and score combination methods, though appealing since
they require no labeled data, did not perform well since they do not weight votes
well.

We found that random forests of decision trees was the best combination method.
We hypothesize that this is due to the nature of our task and base systems.
Random forests were able to help tease apart the high-error region (where
anomalies take place). A drawback of random forests as a combination
method is that they require labeled data. However, experiments reveal
empirically that a relatively small amount of data is sufficient and the 
amount might be able to be further reduced through specific selection criteria.

\section*{Acknowledgments}
This material is based upon work supported, in whole or
in part, with funding from the United States Government.
Any opinions, findings and conclusions, or
recommendations expressed in this material are those of
the author(s) and do not necessarily reflect the views of
the University of Maryland, College Park and/or any
agency or entity of the United States Government.
Nothing in this report is intended to be and shall not be
treated or construed as an endorsement or
recommendation by the University of Maryland, United
States Government, or the authors of the product,
process, or service that is the subject of this report. No
one may use any information contained or based on this
report in advertisements or promotional materials related
to any company product, process, or service or in
support of other commercial purposes.

\bibliographystyle{acl}
\bibliography{paper}

\end{document}